\documentclass[runningheads]{llncs}
\usepackage{graphicx}
\usepackage{algorithm} 
\usepackage{algpseudocode} 
\usepackage{graphicx} 
\usepackage[usenames,dvipsnames]{color}
\usepackage{hyperref}

\begin{document}
\title{Leveraging Human-Machine Interactions for Computer Vision Dataset Quality Enhancement \thanks{*This work has been accepted at the 15th International Conference on Intelligent Human-Computer Interaction (IHCI) 2023.}}
\renewcommand\footnotemark{}
\renewcommand\footnoterule{}
\titlerunning{Human-Machine Dataset Enhancement}

\author{Esla Timothy Anzaku\inst{1,2,3} \and
Hyesoo Hong\inst{1}  \and
Jin-Woo Park\inst{1}   \and
Wonjun Yang\inst{1}   \and
Kangmin Kim\inst{1}    \and
JongBum Won\inst{1}   \and
Deshika Vinoshani Kumari Herath\inst{6}   \and
Arnout Van Messem\inst{5}  \and
Wesley De Neve\inst{1,2,3}  
}

\authorrunning{E. T. Anzaku et al.}
%
\institute{Ghent University Global Campus, Incheon, 21985, South Korea \and
Center for Biosystems and Biotech Data Science, Ghent University Global Campus, Incheon, 21985, South Korea \and
IDLab, Ghent University, Technologiepark-Zwijnaarde 126, B-9052 Ghent, Belgium \and 
University of Liège, 4000 Liège, Belgium \and
Mediio, Seoul, South Korea
 \\
}

\maketitle              
\begin{abstract}

Large-scale datasets for single-label multi-class classification, such as \emph{ImageNet-1k}, have been instrumental in advancing deep learning and computer vision. However, a critical and often understudied aspect is the comprehensive quality assessment of these datasets, especially regarding potential multi-label annotation errors. In this paper, we introduce a lightweight, user-friendly, and scalable framework that synergizes human and machine intelligence for efficient dataset validation and quality enhancement. We term this novel framework \emph{Multilabelfy}.
Central to Multilabelfy is an adaptable web-based platform that systematically guides annotators through the re-evaluation process, effectively leveraging human-machine interactions to enhance dataset quality. By using Multilabelfy on the ImageNetV2 dataset, we found that approximately $47.88\%$ of the images contained at least two labels, underscoring the need for more rigorous assessments of such influential datasets. Furthermore, our analysis showed a negative correlation between the number of potential labels per image and model top-1 accuracy, illuminating a crucial factor in model evaluation and selection. Our open-source framework, Multilabelfy, offers a convenient, lightweight solution for dataset enhancement, emphasizing multi-label proportions. This study tackles major challenges in dataset integrity and provides key insights into model performance evaluation. Moreover, it underscores the advantages of integrating human expertise with machine capabilities to produce more robust models and trustworthy data development. The source code for Multilabelfy will be available at \href{https://github.com/esla/Multilabelfy}{https://github.com/esla/Multilabelfy}

\end{abstract}

\section{Introduction}

Deep learning, the engine behind advanced computer vision, has been largely propelled by training on expansive resources like the ImageNet Large Scale Visual Recognition Challenge (ILSVRC) dataset~\cite{Krizhevsky2017ImageNetNetworks}, commonly known as ImageNet-1k. However, recent performance trends in deep neural network (DNN) models trained on these datasets have shown top-1 and top-5 accuracy stagnation across diverse DNN architectures and training techniques, irrespective of model complexity and dataset size~\cite{RossWightman2019PytorchModels,ozbulak2023know}. This performance plateau suggests that we may be nearing the limits of model accuracy with the current ImageNet-1k dataset using the top-1 accuracy.

A potentially overlooked factor contributing to the observed stagnation may be attributed to the inherent multi-label nature of the dataset in question. It is plausible that a substantial proportion of the images in the dataset are related to more than a single ground truth label. However, the dataset only provides labels for a singular ground truth, which may impose limitations~\cite{Beyer2020AreImageNet,Tsipras2020FromBenchmarks,Vasudevan2022WhenImageNet}. This single-label ground truth constraint could inadvertently lead to underestimating the performance of DNN models, particularly when utilizing the top-1 accuracy metric.

Furthermore, the performance of models significantly degrades when assessed on newer but similar datasets, such as ImageNetV2~\cite{Recht2019DoImageNet}. Despite being developed following a similar protocol to the original ImageNet-1k dataset, ImageNetV2 exhibits unexplained accuracy degradation across various models, regardless of model architecture, training dataset size, or other training configurations. While efforts are being made to investigate this degradation~\cite{Engstrom2020IdentifyingReplication,Anzaku2022ADatasets}, we found only one work that partially studied this problem~\cite{Shankar2020EvaluatingImagenet}.

Prior work has acknowledged the need for more accurate dataset labels and has published reassessed labels that reflect the multi-label nature of the ImageNet-1k validation set~\cite{Beyer2020AreImageNet}. However, label reassessment is not a trivial task. It requires considerable resources and expertise, presenting a substantial challenge for smaller research groups. Given the vital role of the validation and test sets in DNN model selection and benchmarking, meticulous analysis of the ImageNet-1k validation set and its replicates remains indispensable. This critical importance highlights the necessity for accessible and effective frameworks to scrutinize and tackle the multi-label nature of computer vision single-label classification datasets. To address this, we propose an accessible and scalable framework, termed Multilabelfy, that combines human and machine intelligence to efficiently validate and improve the quality of computer vision multi-class classification datasets. Multilabelfy comprises four stages: (i) label proposal generation, (ii) human multi-label annotation, (iii) annotation disagreement analysis, and (iv) human annotation refinement. It is designed with two primary objectives: to strategically harness the capabilities of a diverse pool of annotators and to seamlessly blend human expertise with machine intelligence to improve the quality of a dataset. These objectives are made accessible through a user-friendly interface.

This research effort enriches existing literature by offering Multilabelfy for improving the quality of computer vision multi-class classification datasets. Utilizing Multilabelfy, we reassessed the labels for ImageNetV2, revealing that $47.88\%$ of the images in this dataset could have more than one valid label. We also identified other noteworthy dataset issues. Our work accentuates the importance of recognizing and addressing the multi-label nature of ImageNet-1k and its replicates. Our ultimate goal is to contribute towards developing robust DNN models that can effectively generalize beyond their training data.

\section{Related Work}
\subsection{Label Errors}
Label errors have been identified within the test sets of numerous commonly used datasets, including a $6\%$ error rate in the ImageNet-1k validation set~\cite{Northcutt2021PervasiveBenchmarks}. The importance of tackling the issue of label errors in test partitions of datasets was further emphasized. It was found that high-capacity models are prone to mirroring these systematic errors in their predictions, potentially leading to a misrepresentation of real-world performance and distortion in model comparisons. In another work, an extensive examination of $13,450$ images across $269$ categories in the ImageNet-1k validation set, which predominantly includes wild animal species, was conducted~\cite{Luccioni2023BugsBiodiversity}. Through collaboration with ecologists, it was found that many classes were ambiguous or overlapping. An error rate of $12\%$ in image labeling was reported, with some classes being erroneously labeled more than $90\%$ of the time. Our work further accentuates the critical role of addressing label errors in datasets used for model evaluation. It underscores the need for more precise and thorough dataset construction and assessment methodologies.

\subsection{From Single-Label to Multi-Label}

Single-label evaluation has traditionally served as the standard for assessing models on the ImageNet-1k dataset. However, a reassessment of the ImageNet-1k validation ground truth labels revealed that a good proportion of the images could have multiple valid labels, prompting the creation of Reassessed Labels (ReaL)~\cite{Beyer2020AreImageNet}, incorporating these multi-labels. 

In a related study~\cite{Vasudevan2022WhenImageNet}, the remaining errors that models made on the ImageNet-1k dataset were examined, focusing on the multi-label subset of \emph{ReaL}. Nearly half of the perceived errors were identified as alternative valid labels, confirming the multi-label nature of the dataset. However, it was also observed that even the most advanced models still exhibited about $40\%$ of errors readily identifiable by human reviewers.

\subsection{ImageNet-1k Replicates}
When tested on replication datasets like ImageNetV2, DNN models have been observed to demonstrate a significant, yet unexplained, drop in accuracy~\cite{Recht2019DoImageNet}. Despite these replication datasets, including ImageNetV2, being created by following the original datasets' creation protocols closely, the performance decline raises significant questions about the models' generalization capabilities or the integrity of the datasets. The significant performance drop on ImageNetV2, between $11\%$ to $14\%$~\cite{Recht2019DoImageNet}, was based on the conventional approach of evaluating model accuracy using all data points in the test datasets. 

However, it has been argued that the conventional evaluation approach may not fully capture the behavior of DNN models and may set unrealistic expectations about their accuracy~\cite{Engstrom2020IdentifyingReplication,Anzaku2022ADatasets}. A more statistically detailed exploration into this unexpected performance degradation on ImageNetV2 found that standard dataset replication approaches can introduce statistical bias~\cite{Engstrom2020IdentifyingReplication}. After correcting for this bias and remeasuring selection frequencies, the unexplained part of the accuracy drop was reduced to an estimated $3.6\% \pm 1.5\%$, significantly less than the original $11.7\% \pm 1.0\%$ earlier reported in \cite{Recht2019DoImageNet}. An alternative evaluation protocol that leverages subsets of data points based on different criteria, including uncertainty-related information, provides an alternative perspective~\cite{Anzaku2022ADatasets}. Through comprehensive evaluation leveraging the predictive uncertainty of models, the authors found that the degradation in accuracy on ImageNetV2 was not as steep as initially reported, suggesting possible differences in the characteristics of the datasets that warrant further investigation. A closely related research work studies various aspects of the ImageNet-1K and ImageNetV2 datasets using human annotators. Using a sample of $1,000$ images from both datasets, the proportion of images with multiple labels was estimated to be $30.0\%$ and $34.4\%$ for the two datasets, respectively. This information is detailed in Section B.2 of the supplementary material in ~\cite{Shankar2020EvaluatingImagenet}. The cited work suggested that the difference in the multi-label composition between the two datasets could be a possible explanation for the accuracy degradation.

\subsection{Key Modifications to Existing Approaches}
Our research expands upon a previous work~\cite{Beyer2020AreImageNet} with several essential modifications:
\subsubsection{Model Selection for Candidate Label Proposals} In contrast to the original study's use of a hand-annotated sample of 256 images from a 50,000-image dataset to guide the selection of an optimal model ensemble, we built upon their work, utilizing their generated multi-labels and proposed \emph{ReaL accuracy}. We selected the best-performing pre-trained model utilizing the ReaL accuracy metric, designed to evaluate multi-class classification DNN models on a multi-label test dataset. Further details on this process are provided in Section~\ref{subsec:label_proposal_generation}.
\subsubsection{Image Pre-selection for Multi-label Annotation} The original study only utilized an ensemble of pre-trained single-label models to generate eight candidate labels. In contrast, our approach extended the candidate proposals to $20$, thereby decreasing the risk of omitting valid labels and increasing selection accuracy (Section~\ref{subsec:label_proposal_generation}).

\subsubsection{Annotation Refinement} We introduced an additional stage, wherein the top twenty model-proposed labels, alongside all human-selected labels, are presented to an additional pool of experienced annotators for further refinement (Section~\ref{subsec:human_annotation_refinement}).

\subsubsection{Open-source Platform} Recognizing that platforms like Mechanical Turk might be inaccessible or not affordable for some research labs, we developed Multilabelfy, an open-source alternative. This platform allows in-house dataset quality improvement while maintaining a user-friendly interface. \\

Section~\ref{sec:proposed_framework} provides more comprehensive information regarding these contributions.

\section{Proposed Framework}
\label{sec:proposed_framework}
\subsection{Overview}

\begin{figure}[htbp]
\centering
\includegraphics[scale=0.25]{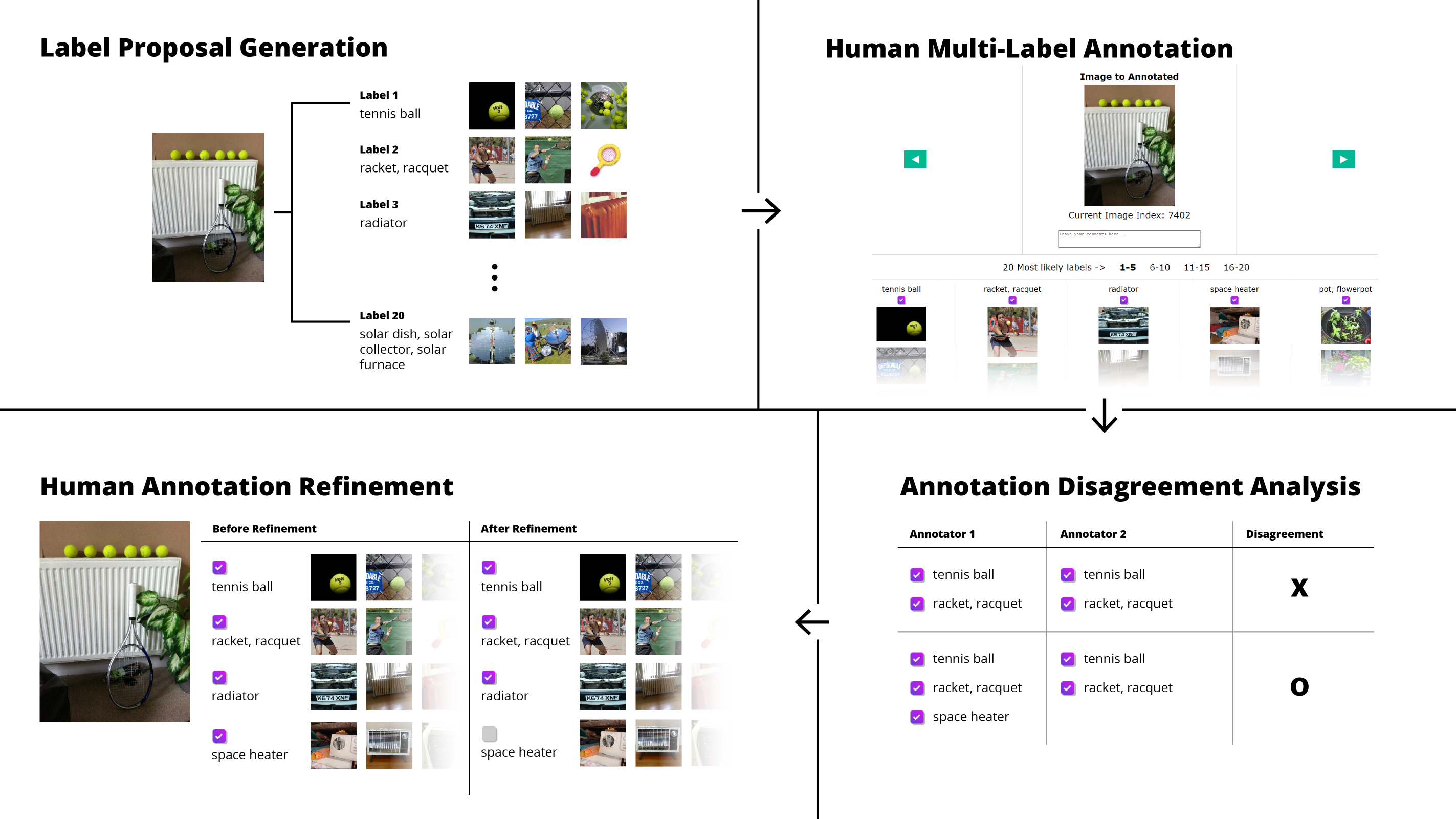} 
\caption{Overview of the proposed framework for enhancing computer vision datasets from single-label to multi-label, enabling a more comprehensive capture of their descriptions.} 
\label{fig:proposed_framework_overview} 
\end{figure}

The proposed multi-label dataset enhancement framework (Figure~\ref{fig:proposed_framework_overview}) comprises four key stages: (i) label proposal generation, (ii) human multi-label annotation, (iii) annotation disagreement analysis, and (iv) human annotation refinement. The label proposal generation and annotation disagreement analysis can be automated using the appropriate algorithms while the human multi-label annotation and human annotation refinement require the involvement of human annotators.

\subsection{Label Proposal Generation}
\label{subsec:label_proposal_generation}

Our qualitative analysis shows that pre-trained models, originally trained on single-label computer vision datasets, can effectively rank the predicted probability vector. This capability is corroborated by the near-perfect top-5 accuracies of state-of-the-art DNN classification models reaching approximately $99\%$~\cite{RossWightman2019PytorchModels}. For model selection, we utilize the \emph{ReaL accuracy} metric, specifically designed to assess the performance of single-label pre-trained DNN models in multi-label scenarios. \emph{Under this metric, an image prediction is considered correct if the prediction belongs to the set of ground truth labels assigned to the image}. The selected model is then used to generate the top-20 label proposals, an increase from the eight proposals presented in previous work, to ensure broader coverage of valid labels. Given the potential for information overload with many label proposals, we designed the annotation user interface to mitigate this concern. Additional details regarding the role of human annotators and the annotation interface are discussed in Section~\ref{subsec:human_multi_label_annotation}.

\begin{figure}[!ht]
\centering
\includegraphics[scale=0.25]{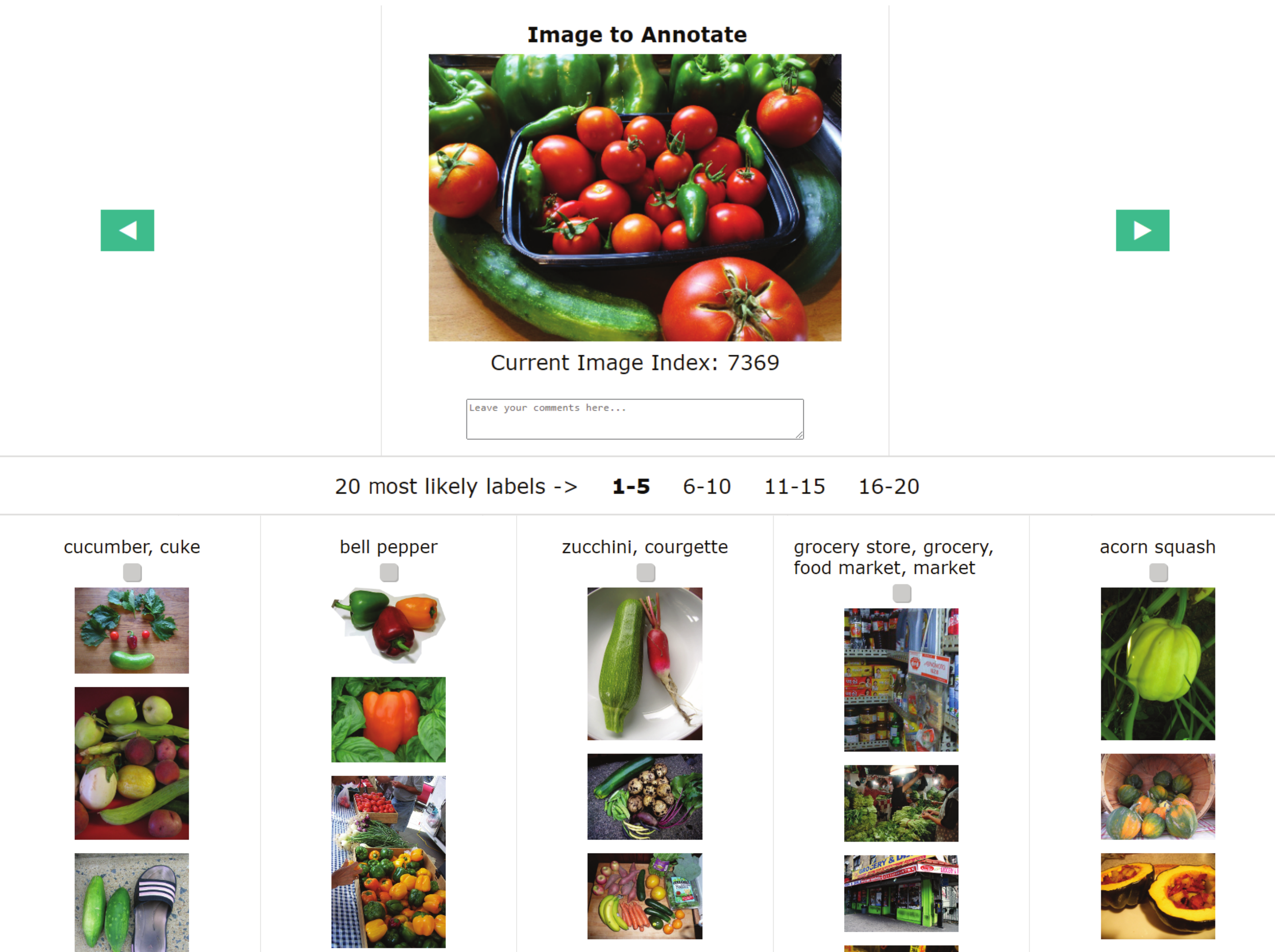} 
\caption{The user interface of the annotation platform. It showcases key features like label presentation in groups of five, a single checkbox per proposed label, scrollable sample images, and click-to-enlarge functionality for detailed inspection of images. These features are designed to streamline the annotation process and efficiently accommodate multi-label data annotation.} 
\label{fig:multi_label_website} 
\end{figure}
\subsection{Multi-Label Annotation by Human Annotators}
\label{subsec:human_multi_label_annotation}

Multilabelfy incorporates a strategically designed web interface to alleviate the workload of human annotators, with a screenshot provided in Figure~\ref{fig:multi_label_website}. This user interface is characterized by several key features engineered to enhance the efficiency and effectiveness of the annotation process. It facilitates the display of label names, their corresponding synonyms, and representative images from the pool of twenty potential labels systematically organized into four subgroups of five labels each. In the event that the initial group does not sufficiently encompass all visible objects, annotators have the option to navigate to other label groups. 

The design also incorporates a streamlined selection process facilitated by a singular checkbox assigned to each proposed label. Moreover, ten exemplar images are presented in a scrollable format for each proposed label, providing a comprehensive view without overwhelming the annotator. Further attention to detail is reflected in the feature that allows images to be clicked on, enabling annotators to inspect these images at their original resolution. These elements combined optimize the multi-label annotation process, yielding higher accuracy and efficiency.

\subsection{Annotation Disagreement Analysis}
\label{subsec:annotation_disagreement_analysis}

Single-label multi-class classification computer vision datasets often comprise images featuring multiple objects. However, a prior research work~\cite{Tsipras2020FromBenchmarks} estimated that about $80\%$ of the ImageNet-1k images contain a single object. We also expect some images with multiple labels to pose no challenges to the annotators. Considering the aforementioned observations, our framework seeks to effectively exclude such images from the pool intended for further refinement. We target images that require additional human annotation refinement during the \emph{annotation disagreement analysis} stage, as depicted in Figure~\ref{fig:proposed_framework_overview}. Images are selected for further annotation refinement if the labels generated by human annotators, as discussed in Section~\ref{subsec:human_multi_label_annotation}, fail to meet a predefined annotation agreement condition. This annotation agreement condition requires: \emph{complete consistency across all labels identified by human annotators for a particular image and the inclusion of the originally provided ground truth label within the array of labels selected by the annotators}. This strategic condition facilitates focused refinement of annotations for the subset of images that pose more significant challenges to annotators. As a result, we minimize the misuse of annotators' time and provide an avenue for a more detailed examination of the more complex images, ultimately fostering a more thoroughly annotated dataset.

\subsection{Refinement of Human Annotation}
\label{subsec:human_annotation_refinement}

This stage follows the process described in Section~\ref{subsec:human_multi_label_annotation} but with some critical distinctions. In the stage described in Section~\ref{subsec:human_multi_label_annotation}, annotators with varying degrees of experience with the dataset contribute to the labeling. However, the refinement stage exclusively engages more experienced annotators. These experienced annotators are provided with the labels previously selected, which are pre-checked for the annotators to review: uncheck (to correct) or check additional missing labels. Furthermore, the annotators are instructed to document any changes they make to the labels using the comments section of the web interface. This provision ensures that a clear record is maintained for each correction, which can be invaluable in resolving potential discrepancies in the annotations. It is important to note that these annotators have undergone several tutorial sessions on the label issues of the ImageNet-1k dataset. Additionally, they reviewed and summarized related literature to ensure that they are aware of the nuanced issues that are encountered when annotating images into $1,000$ categories, especially within the fine-grained categories.

\section{Results}

\subsection{Experimental Setup}
Our goal is to re-assess the labels for the ImageNetV2 dataset to accommodate and account for its multi-label nature. The four stages in Section~\ref{sec:proposed_framework} were carefully followed. In the label proposal generation stage, the EVA-02~\cite{Fang2023Eva-02:Genesis} model was used to generate the proposal. It is one of the top performing models ($90.05\%$ top-1 accuracy~\cite{RossWightman2019PytorchModels}) on the ImageNet-1k dataset; additional details of the model can be found in the cited paper. Subsequently, in the human multi-label annotation stage, the $10,000$ images of the ImageNetV2 dataset were partitioned into seven batches, and each batch was assigned to two human annotators. 
\begin{figure}[htbp]
\centering
\includegraphics[width=\textwidth]{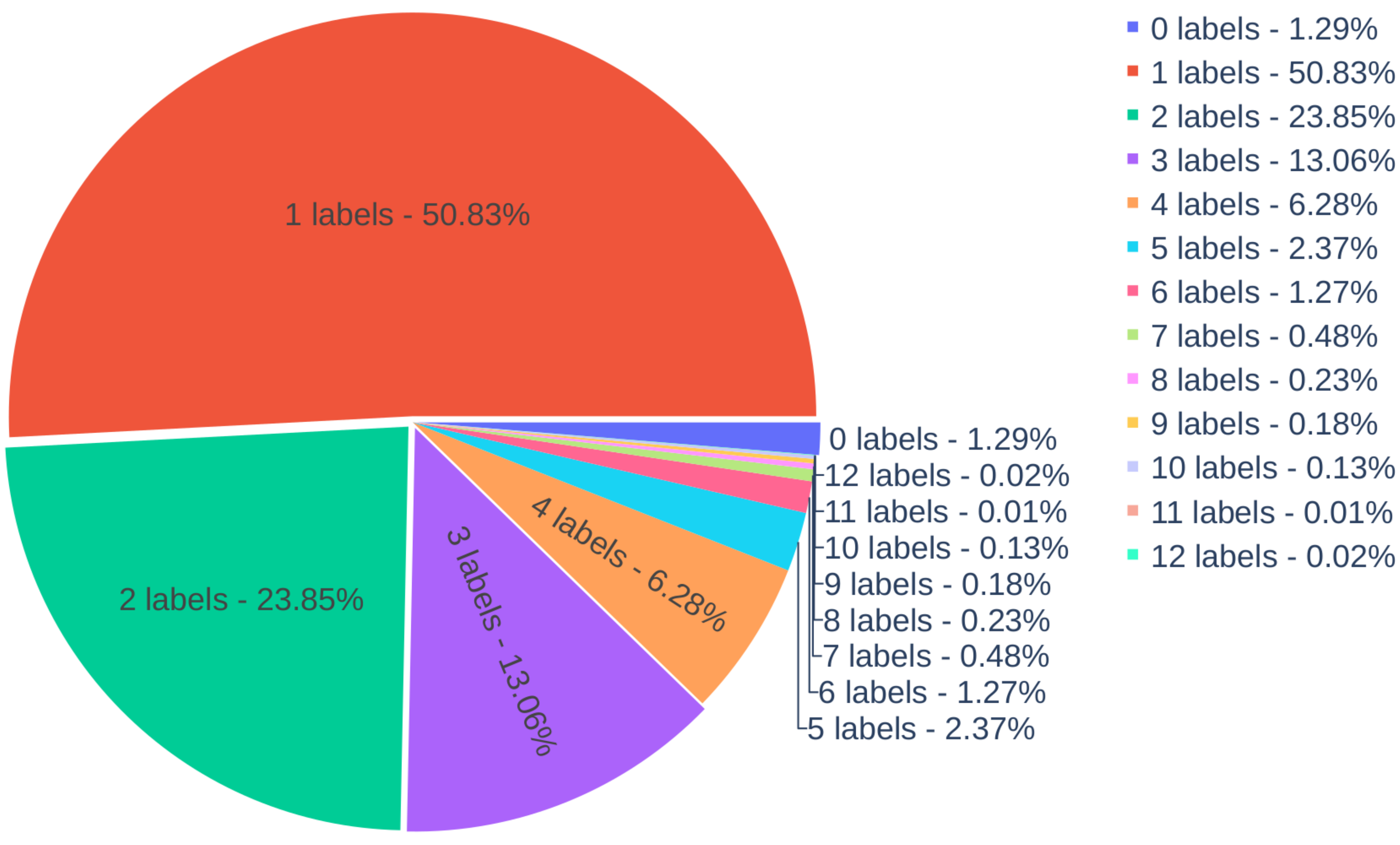} 
\caption{The distribution of images based on the number of labels assigned to them during our annotation process.}
\label{fig:multilabel_pie_chart} 
\end{figure}
Fourteen human annotators having varying experience levels with the ImageNet-1k dataset and computer vision in general, participated during this stage. Upon the annotation disagreement analysis (detailed in Section~\ref{subsec:annotation_disagreement_analysis}), the annotations for $6,425$ of the $10,000$ images fulfilled our disagreement criteria and were selected for subsequent refinement by five more experienced annotators, four of whom were previously referenced among the group of fourteen. Each annotator refined the annotations for $1,285$ images. The refined annotations were then used to generate the results presented and discussed in the following sections.

\subsection{The Extent of ImageNetV2 Multi-Labeledness}
\label{the_extent_of_imagenetv2_multilabeledness}

Here, we provide visual statistics summarizing the multi-label nature of the labels we generated for the ImageNetV2 dataset. Specifically, we show what percentage of the dataset contains which label count, i.e., the number of ground truth labels assigned to an image. As shown in the pie chart of Figure~\ref{fig:multilabel_pie_chart}, the annotation process could not find labels for $1.29\%$ of the images. Moreover, $50.83\%$ of the images contain one label, $23.85\%$ contain two labels, and $24.03\%$ contain more than two labels.

\subsection{Re-Evaluation of Models on ImageNetV2 Improved Labels}

\subsubsection{Top-1 Accuracy versus ReaL Accuracy}

\begin{figure}[htbp]
    \centering
    \includegraphics[scale=0.6]{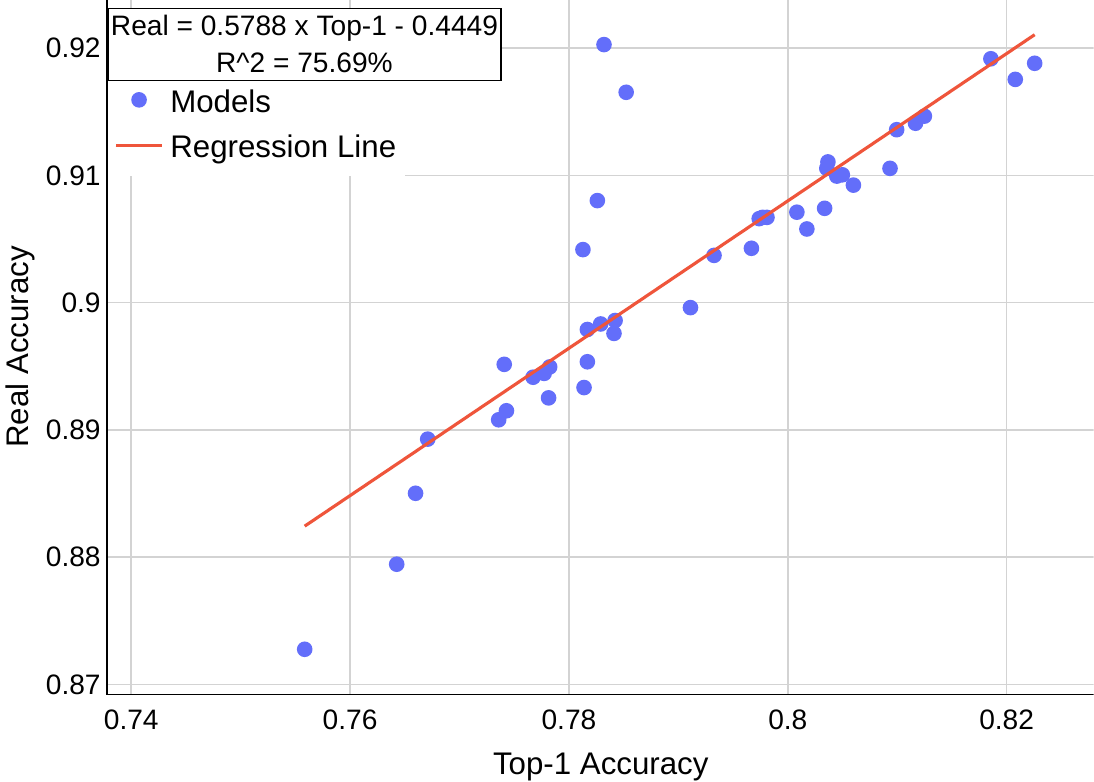}
    \caption{Scatterplot of ReaL accuracy versus top-1 accuracy for $57$ top-performing DNN models, pre-trained either exclusively on the ImageNet-1k dataset or additionally on external datasets.}
    \label{fig:regression_plot}
\end{figure}

We provide a Scatterplot to understand the relationship between ReaL and top-1 accuracy on our generated labels (Figure~\ref{fig:regression_plot}). Each dot in the plot represents a pre-trained model, and $57$ models were evaluated on the ImageNet-1k validation set and ImageNetV2. These models are sourced from a publicly available GitHub repository~\cite{RossWightman2019PytorchModels} and represent state-of-the-art models pre-trained either exclusively on the ImageNet-1k dataset, or on additional external data.  Details of these models can be found together with the paper's code at \href{https://github.com/esla/Multilabelfy}{https://github.com/esla/Multilabelfy}
The regression analysis indicates a significant correlation between the two metrics. Specifically, for every percentage point increase in top-1 accuracy, the ReaL accuracy rises by approximately $0.5788$ percentage points. The coefficient of determination, $R^2$, is $75.69\%$, suggesting that $75.69\%$ of the variation in ReaL accuracy is explained by its linear relationship with top-1 accuracy. This result reflects a consistent positive relationship: as the top-1 accuracy of models improves, there is a proportional increase in ReaL accuracy. It is worth noting that four models visibly diverge from the regression line; these models merit additional scrutiny to identify potential model-specific quirks or underlying reasons for their divergence. A detailed investigation of these models will be addressed in future work.

\subsubsection{Visual Statistics of Top-1 Accuracy versus Image Count}

\begin{figure}[t]
    \centering
    \includegraphics[scale=0.1]{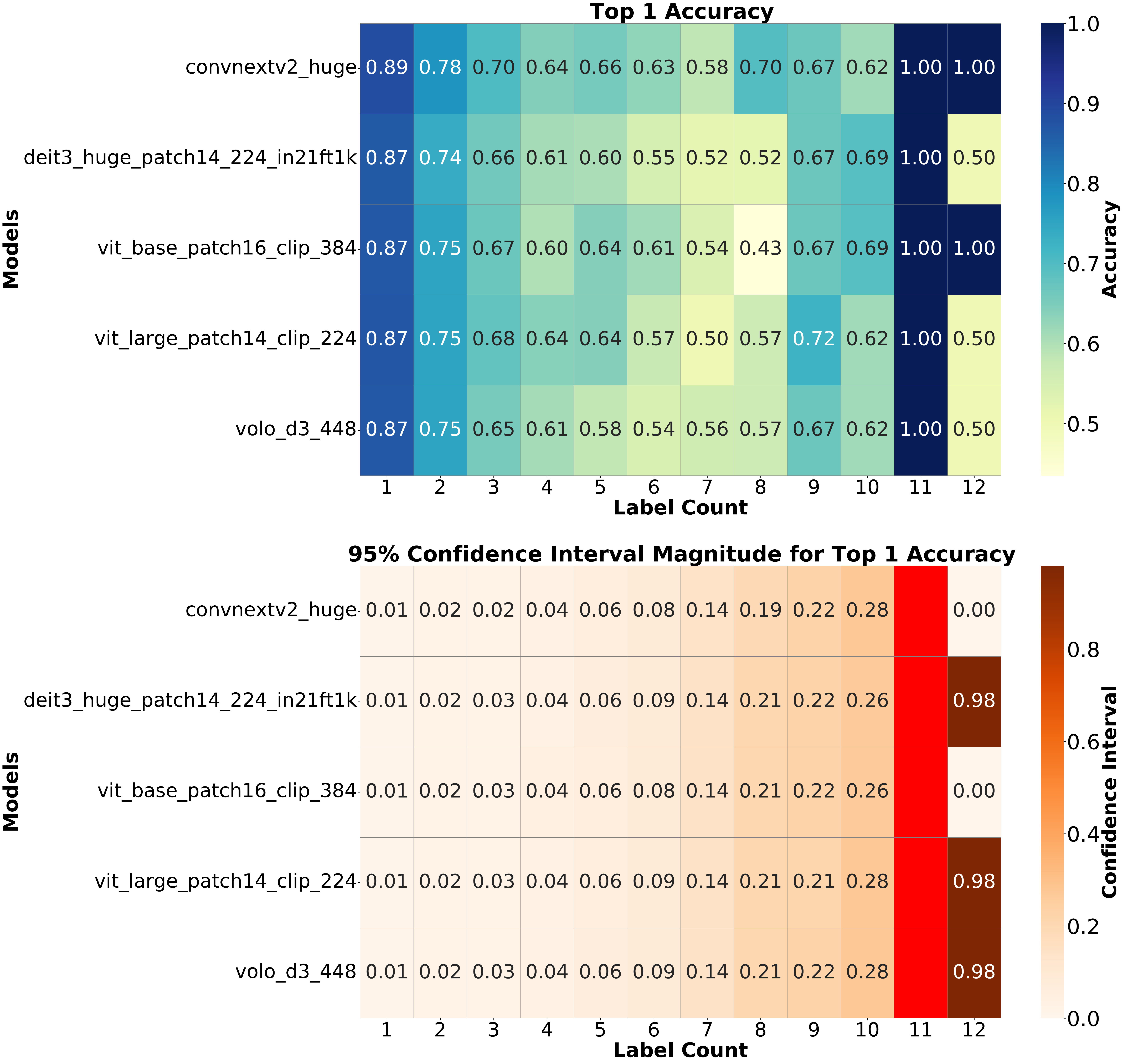}
    \caption{Heatmaps displaying top-1 accuracy (top) for five randomly selected models evaluated on our multi-labeled ImageNetV2 dataset, and the half-width of the $95\%$ confidence interval (bottom) associated with these accuracies. Red cells without numbers represent NaN values due to sets with one or no images for a given label count.
    }

    \label{fig:heatmaps}
\end{figure}

 We investigate the relationship between top-1 accuracy and the variability in image label assignments using heatmaps (Figure~\ref{fig:heatmaps}). While we presented the results for $57$ models in Section~\ref{the_extent_of_imagenetv2_multilabeledness}, for visual brevity, we randomly selected $5$ models for the heatmaps. We determine top-1 accuracy using ground truth labels from the ImageNetV2 dataset, comparing them with our multi-label annotations. To this end, we employ a heatmap (Figure~\ref{fig:heatmaps}, top) that presents top-1 accuracies for each evaluated model across different \emph{label count} categories. While this heatmap is informative, it does not factor in the variability stemming from different sample sizes across label counts. For instance, images with a single label may be more prevalent than those with multiple labels, potentially leading to biases in accuracy measurements.

To enhance our understanding of accuracy computations and account for inherent uncertainties, we incorporate a secondary heatmap as shown in Figure~\ref{fig:heatmaps}, bottom. The margin of error related to the top-1 accuracy is denoted as \( U(i,j) \) and is determined using the following formula: $U(i,j) = 1.96 \times \sigma(i,j)/\sqrt{n}$.
Here, $\sigma(i,j)$ stands for the standard deviation stemming from the binary outcomes of individual predictions for a specific \emph{model} and \emph{label count}. This standard deviation for a binary variable is expressed as $\sigma(i,j) = \sqrt{p(1-p)/n}$, where $p$ represents the proportion of correct predictions. The variable $n$ symbolizes the number of observations for the considered model-label count pairing. This margin of error, corresponding to half the width of the $95\%$ confidence interval, offers a gauge of uncertainty for each model-label count combination. Differences in sample sizes across subsets can lead to variations in the width of the confidence interval. This variance emphasizes the significance of jointly considering both accuracy and its associated uncertainty when interpreting model performance across different label counts.

In our analysis, while results for only five models are presented for clarity, the observations are representative of numerous other models evaluated. A notable observation is that models consistently exhibit higher top-1 accuracy for images associated with a single label. However, as the number of potential labels expands, a discernible decrease in accuracy is evident. This pattern potentially indicates that models might be predicting alternative valid labels, and the top-1 accuracy metric penalizes them for such predictions. Such a negative correlation warrants attention, as it hints at the possibility of underestimating model performance due to potentially skewed dataset assumptions.

\subsection{Analysis of Images with Zero Labels}
\label{subsec:no_labels_analysis}

\begin{figure}[!ht]
    \centering
    \includegraphics[scale=0.6]{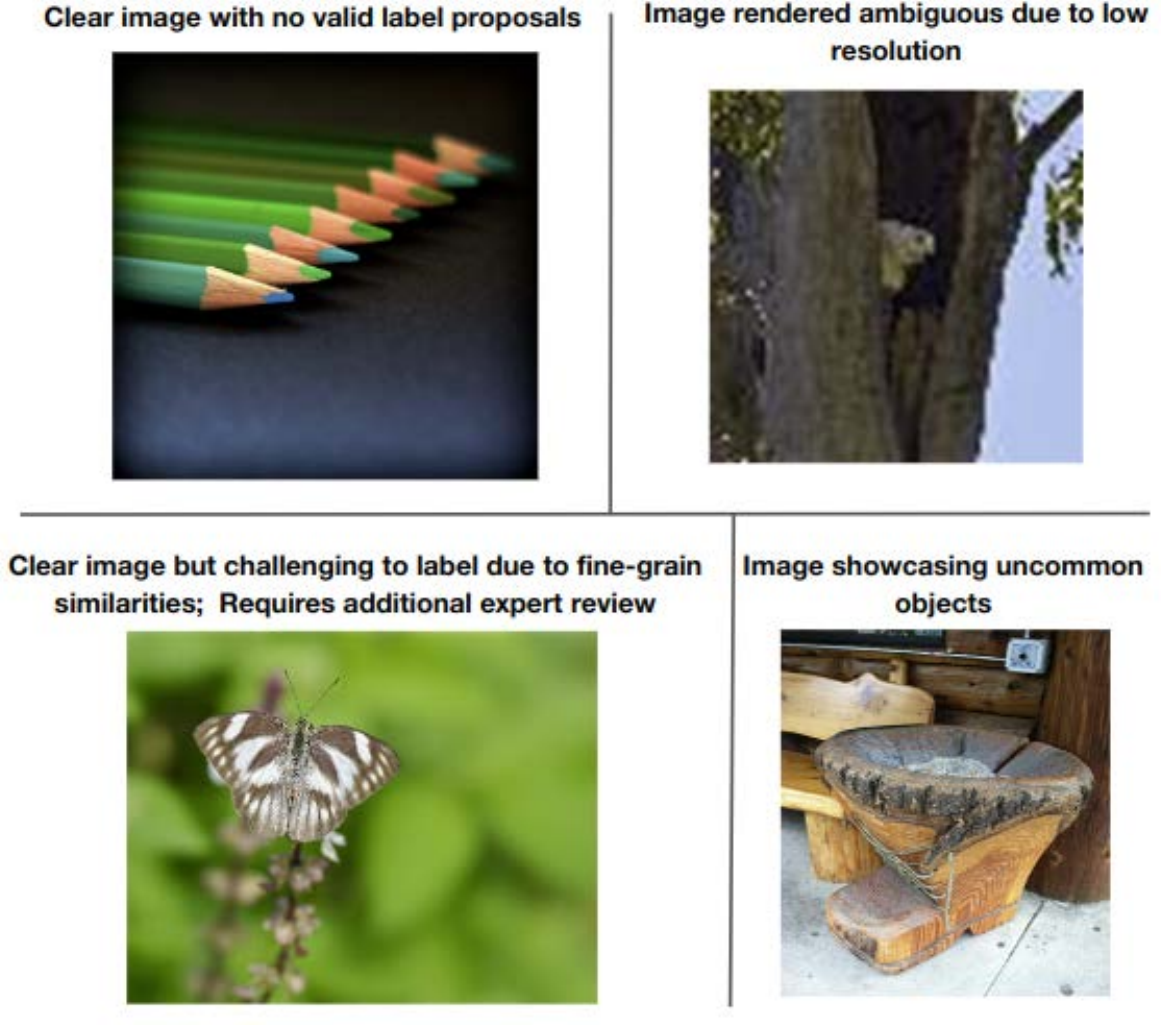}
    \caption{Example images where annotators did not find matching labels from the $20$ proposed labels. The images are categorized based on possible explanations for not finding matching labels in the labels proposed (see Section~\ref{subsec:no_labels_analysis}).}
    \label{fig:no_labels_four_categories}
\end{figure}

\begin{figure}[!htb]
    \centering
    \includegraphics[scale=0.5]{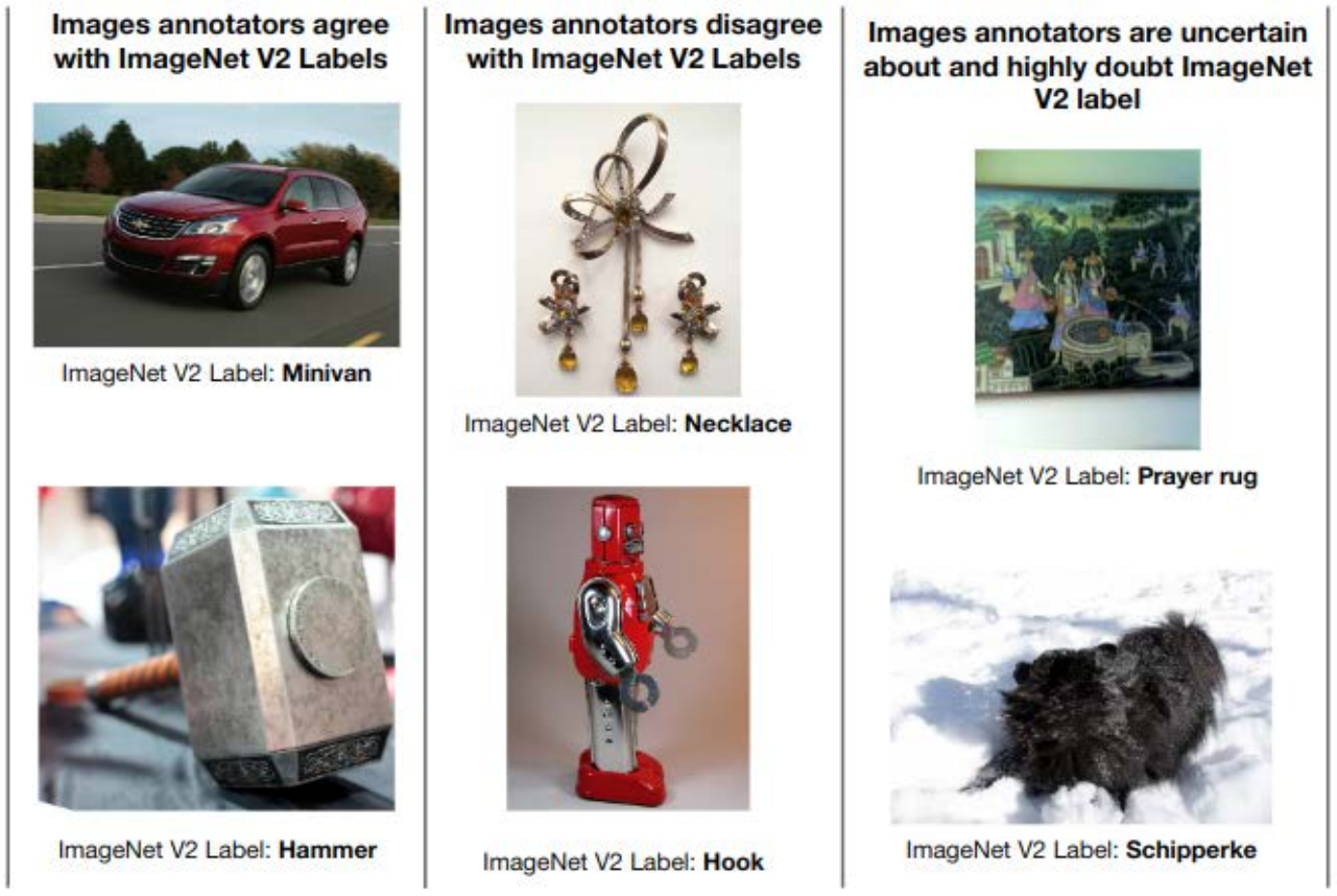}
    \caption{Example images where annotators did not find matching labels from the $20$ proposed labels. The images are categorized based on whether or not our two annotators agree with the provided ImageNetV2 ground truth label (see Section~\ref{subsec:no_labels_analysis}).}
    \label{fig:no_labels_three_categories}
\end{figure}

During our dataset annotation process, despite the meticulous efforts of fifteen annotators, $1.29\%$ ($129$ images) had no labels assigned to them at the completion of human annotation refinement. Consequently, two of the experienced annotators further scrutinized these images. They classified the images without valid annotations into (i) clear images with no valid label proposals ($21.79\%$), (ii) images rendered ambiguous due to low resolution ($10.26\%$), (iii) clear images but challenging to label due to fine-grain similarities, thereby requiring additional expert review ($38.46\%$), and (iv) images showcasing uncommon objects or atypical viewpoints ($29.49\%$). One example from each of these categories is shown in Figure~\ref{fig:no_labels_four_categories}.

While our finalized annotations did not provide labels for these images, ground truth labels from the creators of the ImageNetV2 dataset existed for reference. Using these, the annotators further categorized the images based on their alignment with the ImageNetV2 ground truth as (i) those they agree with ($26.92\%$), (ii) those they disagree with ($19.23\%$), and (iii) those they remain uncertain about and highly doubt ($53.85\%$). Examples of this type of categorization are provided in Figure~\ref{fig:no_labels_three_categories}.

\section{Conclusions}

Single-label multi-class classification datasets like ImageNet-1k are crucial for advancing deep learning in computer vision. However, as the demand for reliable DNN models grows, it is vital to examine these datasets for biases that could impede progress. We provide a practical framework for smaller research groups to enhance the quality of multi-class classification datasets, especially those that could contain multi-labeled images. Furthermore, we introduce new labels for the ImageNetV2 dataset to account for its multi-label nature. The purpose of our dataset enhancement platform and the provided multi-labels for ImageNetV2 is to facilitate research on the performance degradation of ImageNet-1k-trained DNN models on the ImageNetV2 dataset. Interestingly, only about half of the $10,000$ images in the ImageNetV2 dataset can be confidently categorized as having a single label, thereby underscoring the need for further investigation into the impact of the multi-labeled images on ImageNet-based benchmarks and their potential implications for downstream utilization. Such research endeavors will help us better understand how models perform on complex vision datasets.

\subsubsection{Acknowledgment}
This research was supported by Ghent University Global Campus (GUGC) in Korea. This research was also supported under the National Research Foundation of Korea (NRF), (2020K1A3A1A68093469), funded by the Korean Ministry of Science and ICT (MSIT).
We want to specifically thank the following people for their contribution to the annotation process: Gayoung Lee, Gyubin Lee, Herim Lee, Hyesoo Hong, Jihyung Yoo, Jin-Woo Park, Kangmin Kim, Jihyung Yoo, Jongbum Won, Sohee Lee, Sohn Yerim, Taeyoung Choi, Younghyun Kim, Yujin Cho, and Wonjun Yang.


\end{document}